\documentclass[11pt]{article}
\usepackage{eacl2017}
\usepackage{times}
\usepackage{url}
\usepackage{latexsym}
\usepackage{amsthm}
\usepackage{multirow}
\usepackage{tikz}
\usetikzlibrary{positioning,calc,fit}
\usepackage{tkz-graph}
\usepackage{caption}
\usepackage{amsmath}
\usepackage{pgfplots}

\newcommand{\GG}[1]{}
\eaclfinalcopy 

\theoremstyle{definition}
\newtheorem{exmp}{Example}[section]

\title{Use Generalized Representations, But Do Not Forget Surface Features}

\author{Nafise Sadat Moosavi \and Michael Strube\\
  Heidelberg Institute for Theoretical Studies gGmbH \\
  Schloss-Wolfsbrunnenweg 35 \\
  69118 Heidelberg, Germany \\
  {\small \url{{nafise.moosavi|michael.strube}@h-its.org}}}

\date{}

\begin{document}
\maketitle
\begin{abstract}
Only a year ago, all state-of-the-art coreference resolvers were using an extensive amount of surface features.
Recently, there was a paradigm shift towards using word embeddings and deep neural networks, where the use of surface features is very limited. 
In this paper, we show that a simple SVM model with surface features outperforms more complex neural models for detecting anaphoric mentions.
Our analysis suggests that using generalized representations and surface features have different strength that should be both taken into account 
for improving coreference resolution.   
\end{abstract}

\section{Introduction}
\label{sect:intro}
Coreference resolution is the task of finding different mentions that refer to the same entity in a given text.
Anaphoricity detection is an important step for coreference resolution.
An anaphoricity detection module discriminates mentions that are coreferent with one of the previous mentions.
If a system recognizes mention $m$ as non-anaphoric,
it does not need to make any coreferent links for the pairs in which $m$ is the anaphor.

The current state-of-the-art coreference resolvers \cite{wiseman16,clark16a,clark16b}, as well as their anaphoricity detection modules, 
use deep neural networks, word embeddings and a small set of features describing surface properties of mentions.
While it is shown that this small set of features has significant impact on the overall performance \cite{clark16a},
their use is very limited in the state-of-the-art systems in comparison to the embedding features.

In this paper, we first introduce a new neural model for anaphoricity detection that considerably outperforms 
the anaphoricity detection of the state-of-the-art coreference resolver, i.e. deep-coref introduced by \newcite{clark16a}.
However, we show that a simple SVM model that is adapted from our coreferent mention detection approach \cite{moosavi16a}, significantly outperforms the more complex neural models.
We show that the SVM model also generalizes better than the neural model on a new domain other than the CoNLL dataset. 

\section{Discriminating Mentions for Coreference Resolution}
\label{sect:mention}
The recognition of various categories of mentions can be beneficial for coreference resolution.
The detection of the following categories is most common in the literature:
(1) non-referential, (2) discourse-old, and (3) coreferent mentions.
One can also discriminate other categories of mentions 
like mentions that are unlikely to be antecedents or discourse-new mentions \cite{uryupina09}.
However, they are not common in comparison to the above categories.
\subsection{Non-Referential Mentions}
\emph{Non-referential} mentions do not refer to an entity.
These mentions only fill a syntactic position. 
For instance, ``it'' in ``it is raining'' is a non-referential mention.
The approaches proposed by \newcite{evansrichard01}, \newcite{mueller06a}, \newcite{bergsma08}, \newcite{bergsma11} 
are examples of detecting non-referential cases of the pronoun \emph{it}. 
\newcite{byron04} present a more general approach for detecting non-referential noun phrases.
\subsection{Discourse-Old Mentions}
\label{ssect:anaphoricity}
Each mention can be assessed from the point of view of the discourse model \cite{prince92}.
According to the discourse model, a mention may be new, old or inferable.
Mentions which introduce a new entity into the discourse are \emph{discourse-new} mentions.
A discourse-new mention may be a singleton or it may be the first mention of a coreference chain.
For instance, The first ``Plato'' in Example~\ref{anaphoric} is a \emph{discourse-new} mention. 
\begin{exmp}
 \label{anaphoric} 
 \emph{Plato} was a philosopher in Classical Greece. \emph{This philosopher} is the founder of the Academy in Athens. 
  \emph{Plato} died at the age of 81.
\end{exmp}
A \emph{discourse-old} mention refers to an entity that is already evoked in the discourse.
Except for first mentions of coreference chains, other coreferent mentions are \emph{discourse-old}.
For instance, ``this philosopher'' and the second ``Plato'' in Example~\ref{anaphoric} are \emph{discourse-old} mentions.

A mention is \emph{inferable} if the hearer can infer the identity of the mention from another entity that has already been evoked in the discourse.
``the windows'' in Example~\ref{inferable} is an \emph{inferable} mention.
\begin{exmp}
 \label{inferable}
 I walked into \emph{the room}. \emph{The windows} were all open.
\end{exmp}

The detection of discourse-old mentions is commonly referred to as  \emph{anaphoricity detection} 
(e.g.\ \newcite{zhouguodong09}, \newcite{ng09}, \newcite{wiseman15}, \newcite{lassalle15}, inter alia)
while the task of anaphoric mention detection, based on its original definition, is of no use for coreference resolution.
Mentions whose interpretations do not depend on previous mentions are called \emph{non-anaphoric} mentions \cite{vandeemter00}.  
For example, both "Plato"s in Example~\ref{anaphoric} are non-anaphoric.


For consistency with the coreference literature, 
we refer to the task of discourse-old mention detection as anaphoricity detection.

Currently, all the state-of-the-art coreference resolvers learn anaphoricity detection jointly with coreference resolution \cite{wiseman15,wiseman16,clark16a}.
The approaches proposed by \newcite{ng02c}, \newcite{ng04}, \newcite{ng09}, \newcite{zhouguodong09}, \newcite{uryupina09} 
are examples of independent anaphoricity detection approaches.  
\subsection{Coreferent Mentions}
\label{ssect:coreferent}
\newcite{demarneffe15} discriminate mentions as coreferent vs.\ non-coreferent. 
Coreferent mentions are those mentions that appear in a coreference chain.
A non-coreferent mention therefore can be a non-referential noun phrase or a referential noun phrase whose entity is only mentioned once (i.e. singleton).
The proposed approaches of \newcite{recasens13a}, \newcite{demarneffe15}, and \newcite{moosavi16a} discriminate mentions for coreference resolution this way.
\section{Anaphoricity Detection Models}
Anaphoricity detection is the most common approach for discriminating mentions for a coreference resolver.
All of the state-of-the-art coreference resolvers use anaphoricity detection.
In this paper, we compare three different anaphoricity detection approaches: two approaches using neural networks and word embeddings, 
and one using an SVM model and surface features.
\newcite{clark16a} introduce the first neural model.
Since \newcite{clark16a} train their anaphoricity model jointly with the coreference model, 
we refer to this model as the joint model.
We introduce a new anaphoricity detection model as the second neural model using a Long-Short Term Memory (LSTM) network \cite{hochreiter97}.
The third approach is adapted from our state-of-the-art coreferent mention detection \cite{moosavi16a}.
\subsection{Joint Model}
As one of the neural models for anaphoricity detection,
we consider the anaphoricity module of deep-coref\footnote{Available at \url{https://github.com/clarkkev/deep-coref}}, 
the state-of-the-art coreference resolution system introduced by \newcite{clark16a}.
This model has three layers for encoding different types of information regarding a mention.
The first layer encodes the word embeddings of the head, first, last, two previous/following words, and the syntactic parent of the mention.
The second layer encodes the averaged word embeddings of the five previous/following words, all words of the mention, sentence words, and document words.
The third layer encodes the following features of a mention: type, length, position and whether it is embedded in another mention.
The outputs of these three layers are combined into one vector and then get passed through a network with two hidden layers.
This anaphoricity model is trained jointly with the deep-coref coreference model.
\subsection{LSTM Model}
\label{sect:lstm}
In this section we propose a new neural model for anaphoricity detection.
Apart from the properties of the mention itself, 
we consider a limited number of surrounding words.
We first generalize the context of a mention
by removing the mention from the context and replacing it with a special placeholder.
In our experiments, we consider the 10 previous and following words of a mention. 
We concatenate the mention tokens and the head token to the generalized word sequence.
We separate the head and mention tokens in the concatenated sequence using two different placeholders.

The word embeddings of the above sequence are encoded using a bidirectional LSTM.
LSTMs show convincing results on generating meaningful representations  
for various NLP tasks (e.g.\ \newcite{sutskever14} and \newcite{vinyals14}).

We also incorporate a set of surface features
that contains (1) mention type (proper, nominal (definite, indefinite), pronouns (\emph{he, I, it, she, they, we, you})),
(2) string match in the text, (3) string match in the previous context,
(4) head match in the text, (5) head match in the previous context, 
(6) contains tokens of another mention, (7) contains tokens of a previous mention,
(8) contained in another mention, (9) contained in a previous mention, and (10) embedded in another mention. 
These features are concatenated with the output of the bidirectional LSTM and get passed
through one more layer that generates the output.

We also experiment with a more complex model including two different LSTMs for 
encoding mentions and their surrounding words.
We consider longer sequences of previous words 
and an attention mechanism for processing the long sequence.
However, the performance did not improve upon the LSTM model while it considerably increased the training time. 
\subsubsection{Implementation Details}
Hyperparameters are tuned on the CoNLL 2012 development set.
We minimize the cross entropy loss using gradient-based optimization and the Adam update rule \cite{kingma14}. 
We use minibatches of size 50.
A dropout \cite{hinton12} with a rate of 0.3 is applied to the output of LSTM. 
We initialize the embeddings with the 300-dimensional Glove embeddings \cite{glove}.
The size of LSTM's hidden layer is set to 128.
The model is trained in only one epoch.

\subsection{SVM Model}
\label{sect:svm}
Our SVM model introduced in \newcite{moosavi16a}, achieves state-of-the-art results for coreferent mention detection.
This model uses the following set of features: 
lemmas and POS tags of all words of a mention, lemmas and POS tags of the two previous/following words, 
mention string, mention length, mention type (proper, nominal, pronoun, list), string match in the text, and head match in the text.
We use a similar SVM model for anaphoricity detection.
In addition to the features we used for coreferent mention detection, we also add the following features for anaphoricity detection:
string match in the previous context, head match in the previous context, mention words are contained in another mention, 
mention words are contained in a previous mention, mention contains words of another mention, mention contains words of a previous mention.
Similar to \newcite{moosavi16a}, we use an anchored SVM \cite{goldberg07svm} with a polynomial kernel of degree two and remove feature-values that occur less than 10 times.
The use of an anchored SVM with pruning helps the model to generalize better on new domains \cite{goldberg09}.

\section{Performance Evaluation}
\label{conll}
We evaluate the anaphoricity models on the CoNLL 2012 dataset.
It is worth noting that all of the examined anaphoricity detectors in this section use the same mention detection module
and results are reported using system detected mentions.
The performance of the mention detection module is of crucial importance for anaphoricity detection.
Therefore, it is important that the compared anaphoricity detectors use the same mention detection.
\begin{table}[!htbp]
\begin{center}
 \small
 \setlength\tabcolsep{5pt}

 \begin{tabular}{@{}l|lll|lll@{}}
\multicolumn{1}{c}{} & \multicolumn{3}{c@{}}{Non-Anaphoric} & \multicolumn{3}{c}{Anaphoric} \\ \hline
\multicolumn{1}{c}{} & R & P & F1 & R & P &  F1 \\ \hline
   \scriptsize joint \small &- & -& -& $81.81$ & $77.18$ & $79.43$ \\
   \scriptsize LSTM \small & $90.71$ & $92.64$ & $91.66$ & $85.00$ & $81.48$ & $83.20$ \\
   \scriptsize LSTM$^*$ \small & $90.51$ & $87.31$ & $88.88$ & $72.64$ & $78.64$ & $75.52$ \\
   
   \scriptsize {SVM} \small & $92.42$ & $92.61$ & $92.51$ & $84.66$ & $84.30$ & $84.48$  \\ \hline
\end{tabular}
\end{center}
\caption{Results on the CoNLL 2012 test set.\label{table:conll}}
\end{table}

The LSTM model that is described in Section~\ref{sect:lstm} is denoted as \emph{LSTM} in Table~\ref{table:conll}.
In order to investigate the effect of the used surface features, 
we also report the results of the LSTM model without using these features (\emph{LSTM$^*$}).

The following observations can be drawn from the results of Table~\ref{table:conll}:
(1) our LSTM model outperforms the joint model while using less features and being trained independently,
(2) the results of the LSTM$^*$ model is considerably lower than those of LSTM, especially for recognizing anaphoric mentions, 
and (3) the simple SVM model outperforms the neural models in detecting both anaphoric and non-anaphoric mentions.
\subsection{Generalization Evaluation}
In order to investigate the generalization on new domains, we evaluate the LSTM and SVM models on the WikiCoref dataset \cite{ghaddar16}.
The WikiCoref dataset is annotated according to the same annotation guideline as that of CoNLL. 
Therefore, it is an appropriate dataset for performing out-of-domain evaluations when CoNLL is used for training.
For the experiments of Table~\ref{table:wikicoref}, all models are trained on the CoNLL 2012 training data and 
tested on the WikiCoref dataset.

The word dictionary that is used for the LSTM model is built based on the CoNLL 2012 training data.
All words that are not included in this dictionary are treated as out of vocabulary words with randomly initialized word embeddings.
We further improve the performance of LSTM on WikiCoref, by adding the words from the WikiCoref dataset into its dictionary.
The LSTM model trained with this extended dictionary is denoted as \emph{LSTM$^\dagger$} in Table~\ref{table:wikicoref}.
\emph{LSTM$^\dagger$} results are still lower than those of the SVM model while SVM does not use any information from the test dataset. 
Pruning rare lexical features from the training data
along the incorporation of part of speech tags, which are far more generalizable than lexical features,
could explain the generalizability of the SVM model on the new domain.
%
\begin{table}[!htbp]
\begin{center}
\small
\setlength\tabcolsep{5pt}
\begin{tabular}{@{}l|lll|lll@{}}
\multicolumn{1}{c}{} & \multicolumn{3}{c@{}}{Non-Anaphoric} & \multicolumn{3}{c}{Anaphoric} \\ \hline
\multicolumn{1}{c}{} & R & P & F1 & R & P &  F1 \\ \hline
   \scriptsize LSTM \small &  $95.53$ & $89.88$ & $92.62$ & $69.50$ & $84.58$ & $76.31$ \\
   \scriptsize  LSTM$^\dagger$ \small & $93.25$ & $92.78$ & $93.01$ & $79.41$ & $80.57$ & $79.99$  \\   
   \scriptsize {SVM} \small & $93.83$ & $93.05$ & $93.43$ & $80.11$ & $82.07$ & $81.08$ \\
   \hline
\end{tabular}
\end{center}
\caption{Results on the WikiCoref dataset.\label{table:wikicoref}}
\end{table}

\section{Analysis Based on Mention Types}
\label{sect:analysis}
We analyze the output of the LSTM and SVM models on the CoNLL 2012 test set 
to see how well they perform for different types of mentions.
As can be seen from Table~\ref{tab:analysys},
there is not much difference between the performance of LSTM and SVM for recognizing anaphoric pronouns.
SVM detects anaphoric proper names better while LSTM is better at recognizing anaphoric common nouns.

We also analyze the output of LSTM$^*$.
As can be seen, the incorporation of surface features does not affect the detection of anaphoric pronouns very much
while it mainly affects the detection of anaphoric proper names by about 24 percent.

%
In order to see whether the same pattern holds for coreference resolution,
we compare the recall and precision errors of the best coreference system that only uses surface features, 
i.e.\ cort \cite{martschat15c} with singleton features \cite{moosavi16a} \footnote{Available at \url{https://github.com/ns-moosavi/cort/tree/singleton_feature}},
and the state-of-the-art deep coreference resolver, i.e.\ deep-coref \cite{clark16a}.
The comparison of the errors for the CoNLL 2012 test set is shown in Table~\ref{table:error}.
We use the error analysis tool of cort introduced by \newcite{martschat14} for the results of Table~\ref{table:error}.
As can be seen from Table~\ref{table:error},
while deep-coref is significantly better than cort for resolving common nouns and specially pronouns,
its result does not go far beyond that of cort when it comes to resolving proper names.

\begin{table}[!htbp]
\begin{center}
 \small
 \setlength\tabcolsep{5pt}
 \begin{tabular}{@{}l|lll|lll@{}}
\multicolumn{1}{c}{} & \multicolumn{6}{c}{Anaphoric} \\ \hline
\multicolumn{1}{c}{} & R & P & F1 & R & P & F1 \\ \hline
 \multicolumn{1}{c}{} &\multicolumn{3}{c}{Proper names}&\multicolumn{3}{c}{Common nouns} \\ \hline
 \scriptsize LSTM \small & $79.49$ & $82.31$ & $80.88$ & $62.96$ & $65.04$ & $63.99$\\ 
 \scriptsize LSTM$^*$ \small& $47.60$ & $70.09$ & $56.69$ & $46.30$ & $57.75$ & $51.40$\\ 
 \scriptsize SVM \small & $83.80$ & $85.71$ & $84.74$& $52.46$ & $71.98$ & $60.69$\\
\hline
    
\multicolumn{1}{c}{} &\multicolumn{3}{c}{Pronouns} &\multicolumn{3}{c}{Other} \\ \hline
   \scriptsize  LSTM \small & $94.67$ & $85.60$ & $89.91$ & $29.11$ & $63.88$ & $40.00$\\ 
   \scriptsize  LSTM$^*$ \small& $92.67$ & $86.01$ & $89.22$ & $10.13$ & $34.78$ & $15.69$\\
   \scriptsize SVM \small & $95.59$ & $86.29$ & $90.71$& $32.91$ & $76.47$ & $46.02$\\

   \hline
   \end{tabular}
\end{center}
\caption{Anaphoricity results for each mention type on the CoNLL 2012 test set.\label{tab:analysys}}
\end{table}
\begin{table}[!htbp]
\begin{center}
 \small
 \begin{tabular}{l|lll}
& Name & Noun & Pronoun \\ \hline
& \multicolumn{3}{c}{\#Recall Errors} \\ \hline
deep-coref & $1110$ & $1499$ & $1537$ \\
cort & $1145$ & $1638$ & $1655$ \\
\hline
& \multicolumn{3}{c}{\#Precision Errors} \\ \hline
deep-coref & $713$ & $672$ & $1162$ \\
cort & $738$ & $747$ & $1736$ \\
   \end{tabular}
\end{center}
\caption{Coreference error analysis.\label{table:error}}
\end{table}

\section{Discussion}
In this paper we analyze the effect of surface features for anaphoricity detection, 
which is a small but an important step for coreference resolution.
Our analysis shows that surface features, as it was known, are important.
Based on our results, the effects of incorporating surface properties and generalized representations are different for different types of mentions.
These results suggest that apart from a unified model, we should consider different models or at least different features for processing 
different types of mentions
and do not put all the burden on a single model to learn the differences.
The works by \newcite{lassalle13} and \newcite{denis08} are examples of models in which distinct models have been used for various types of mentions.
Besides, our analysis  
shows the importance of surface features for proper names.
Word embeddings are very useful for capturing semantic relatedness.
A coreference resolver that uses word embeddings has a great advantage in better resolution of common nouns and pronouns.
However, the use of surface features in current state-of-the-art coreference resolvers is very limited.
Before going towards using more sophisticated knowledge sources, 
there are still easy victories that can be achieved by incorporating more generalizable surface properties, especially for proper names. 

\section*{Acknowledgments} The authors would like to thank Kevin Clark for his help with the deep-coref software
and Mark-Christoph M{\"u}ller for his helpful comments. 
We would also like to thank the four anonymous reviewers for their detailed comments on an earlier draft of the paper. 
This work has been funded by the Klaus Tschira Foundation, Heidelberg,
Germany. The first author has been supported by a Heidelberg Institute for Theoretical Studies
PhD.\ scholarship.

\bibliographystyle{eacl2017}
\bibliography{mybib}
\end{document}